\documentclass{bmvc2k}

\usepackage{lineno,hyperref}
\usepackage{float}
\usepackage{rotating}
\usepackage{array,graphicx} 
\usepackage{booktabs}
\usepackage{pifont}
\usepackage{caption}
\usepackage{subcaption}
\usepackage{amsmath}
\usepackage{multirow}
\usepackage{float}
\usepackage{multirow}

\usepackage{epstopdf}
\usepackage{url}
\usepackage{amsfonts}

\DeclareMathOperator*{\argmin}{arg\,min}

\title{Automatic Recognition of Mammal Genera on Camera-Trap Images using Multi-Layer Robust Principal Component Analysis and Mixture Neural Networks}

\addauthor{Jhony-Heriberto Giraldo-Zuluaga}{jhonygiraldoz@gmail.com}{1}
\addauthor{Augusto Salazar}{augusto.salazar@udea.edu.co}{1}
\addauthor{Alexander Gomez}{alviurlex@gmail.com}{1}
\addauthor{Ang\'elica Diaz-Pulido}{adiaz@humboldt.org.co}{2}

\addinstitution{
Grupo de Investigaci\'on SISTEMIC, Universidad de Antioquia,\\
Medell\'in, Colombia\\
}
\addinstitution{
Instituto de Investigaci\'on de Recursos Biol\'ogicos Alexander von Humboldt,\\
Bogot\'a D.C, Colombia
}

\runninghead{Giraldo-Zuluaga et al.}{Automatic Recognition of Mammal Genera}

\begin{document}

\maketitle

\begin{abstract}

The segmentation and classification of animals from camera-trap images is due to the conditions under which the images are taken, a difficult task. This work presents a method for classifying and segmenting mammal genera from camera-trap images. Our method uses Multi-Layer Robust Principal Component Analysis (RPCA) for segmenting, Convolutional Neural Networks (CNNs) for extracting features, Least Absolute Shrinkage and Selection Operator (LASSO) for selecting features, and Artificial Neural Networks (ANNs) or Support Vector Machines (SVM) for classifying mammal genera present in the Colombian forest. We evaluated our method with the camera-trap images from the Alexander von Humboldt Biological Resources Research Institute. We obtained an accuracy of 92.65\% classifying 8 mammal genera and a False Positive (FP) class, using automatic-segmented images. On the other hand, we reached 90.32\% of accuracy classifying 10 mammal genera, using ground-truth images only. Unlike almost all previous works, we confront the animal segmentation and genera classification in the camera-trap recognition. This method shows a new approach toward a fully-automatic detection of animals from camera-trap images.




\end{abstract}

\section{Introduction}
\label{sec:intro}

Studying and monitoring wildlife can be achieved by means of non-invasive sampling techniques such as the camera trapping approach. This method captures digital images of wild animals, using small devices composed of a digital camera and a passive infrared sensor. Camera trapping helps the biologist to sample animal populations and to observe species for conservation purposes, e.g. delineating species distributions, monitoring animal behavior, and detecting rare species \cite{swanson2015snapshot}.
A genus is a taxonomic category that includes a group of species sharing certain common characteristics. It is important to explain the difference between genus and species, because we made genus recognition, speaking in biological terms. But, we confronted the same species-recognition problem from the computer-science literature.





Although the camera traps should only capture animal images, the method generates a lot of false positive captures (images without animals). For instance, the camera trapping study performed by Diaz-Pulido et al. \cite{diaz2011densidad} where only $1\%$ of that information was valuable, or the Snapshot Serengeti database \cite{swanson2015snapshot} where $26.8\%$ of the images contain animals.As a result, wildlife scientists must analyze thousands of photographs, of which a high percentage does not show wildlife. This problem, albeit very well known in the camera-trap community, is far from being solved. Furthermore, biologists must classify tens of animal species or genera from thousands of images. An automatic segmentation and classification system might accelerate the professional work, allowing the biologists to concentrate in data analysis.

The pattern recognition community has approached the camera-trap recognition problem in two ways: segmentation (detect animals in images and segment them) and species or genera identification (classification). Yu et al. \cite{yu2013automated} confronted species identification classifying $18$ animal species from a camera-trap database taken in Panama. They used dense Scale Invariant Feature Transform and cell-structured Local Binary Pattern as feature extraction, and multi-class SVM to classify the features; they achieved $82\%$ of accuracy. Note that Yu et al. assumed a perfect segmentation algorithm (doing a manual selection) which always resulted in a perfect animal segmentation (rejecting all false positives). Kumar et al. \cite{kumar2015animal} also confronted species identification. They classified $30$ animal species with a human aided segmentation method, and extracting eight Local Fourier Transform for feeding a K-Nearest Neighbor and a Probabilistic Neural Network. They reached an $82.7\%$ of classification accuracy.

Chen et al. \cite{chen2014deep} faced the segmentation and identification using $20$ animal species from a camera-trap database taken in North America. First, they segmented the images with Ensemble Video Object Cut and then classified the segmented images comparing the performance of Bag of Visual Words with a CNN architecture with 3 convolutional and 3 max pooling layers. Although the Chen et al. method was designed removing false positives, the reached performance was low ($38.31\%$ of accuracy) compared with manual segmentation. Gomez et al. confronted the species identification problem in two scenarios. First, they classified $26$ animal species from the Snapshot Serengeti dataset \cite{swanson2015snapshot}, using very deep CNNs such as AlexNet, VGGNet, GoogLeNet and ResNet. They achieved $88.9\%$ of accuracy \cite{gomez2016towards}. Second, Gomez et al. classified between two groups of mammals using deep CNNs on low quality camera-trap images. They achieved $90.35\%$ of accuracy \cite{gomez2016animal}. In both cases, the best accuracies assume a perfect segmentation. In contrast, Giraldo et al. \cite{giraldo2017camera} only faced the segmentation problem. They proposed the Multi-Layer RPCA method in order to solve the segmentation problem on camera-trap images. They reached $75.39\%$ and $73.93\%$ of average f-measure in daytime and nighttime images respectively, but they did not perform species classification.


In this work, unlike almost all previous works, the two camera-trap problems (segmentation and genera classification) are faced. We propose an automatic genera recognition method based on background subtraction techniques and very deep CNNs. Our method is composed of Multi-Layer RPCA segmentation, CNN feature extraction, LASSO selection, and ANN or SVM classification. A comparison between manual and automatic segmentation accuracy as well as multiple CNNs mainstream architectures is done.


The paper is organized as follows: Section \ref{sec:camera-trapImages} describes the camera-trap images. Section \ref{sec:identificationMethod} introduces the identification method. Section \ref{sec:expFram} describes the experimental framework. Section \ref{sec:results} presents the experimental results and the discussion. Finally, Section \ref{sec:conclusions} shows the conclusions and future works.



\section{Camera-trap images}
\label{sec:camera-trapImages} 

\begin{figure}[h]
    \begin{center}
    \begin{subfigure}[b]{0.2\textwidth}
		\centering
		\includegraphics[height=1.8cm,keepaspectratio,]{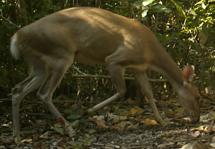}
        \caption{}
        \label{fig:idealImage}
    \end{subfigure}
    \begin{subfigure}[b]{0.2\textwidth}
		\centering
		\includegraphics[height=1.8cm,keepaspectratio,]{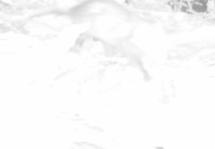}
        \caption{}
        \label{fig:overExposed}
    \end{subfigure}
    \begin{subfigure}[b]{0.2\textwidth}
		\centering
		\includegraphics[height=1.8cm,keepaspectratio,]{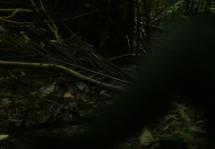}
        \caption{}
        \label{fig:parts}
    \end{subfigure}
    \begin{subfigure}[b]{0.2\textwidth}
		\centering
		\includegraphics[height=1.8cm,keepaspectratio,]{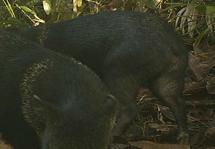}
        \caption{}
        \label{fig:variousAnimal}
    \end{subfigure}
    \begin{subfigure}[b]{0.2\textwidth}
		\centering
		\includegraphics[height=1.8cm,keepaspectratio,]{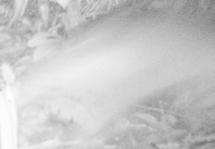}
        \caption{}
        \label{fig:blurred}
    \end{subfigure}
    \begin{subfigure}[b]{0.2\textwidth}
		\centering
		\includegraphics[height=1.8cm,keepaspectratio,]{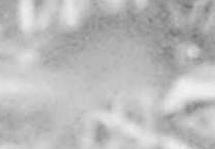}
        \caption{}
        \label{fig:lowResolution}
    \end{subfigure}
    \begin{subfigure}[b]{0.2\textwidth}
		\centering
		\includegraphics[height=1.8cm,keepaspectratio,]{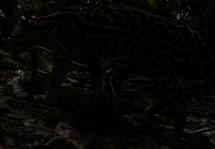}
        \caption{}
        \label{fig:poorIllumination}
    \end{subfigure}
    \begin{subfigure}[b]{0.2\textwidth}
		\centering
		\includegraphics[height=1.8cm,keepaspectratio,]{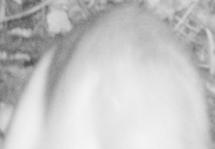}
        \caption{}
        \label{fig:autoOclussion}
    \end{subfigure}
    \begin{subfigure}[b]{0.18\textwidth}
		\centering
		\includegraphics[height=2.2cm,keepaspectratio,]{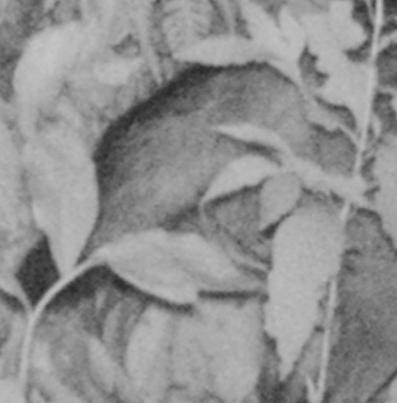}
        \caption{}
        \label{fig:context}
    \end{subfigure}
    \begin{subfigure}[b]{0.18\textwidth}
		\centering
		\includegraphics[height=2.2cm,keepaspectratio,]{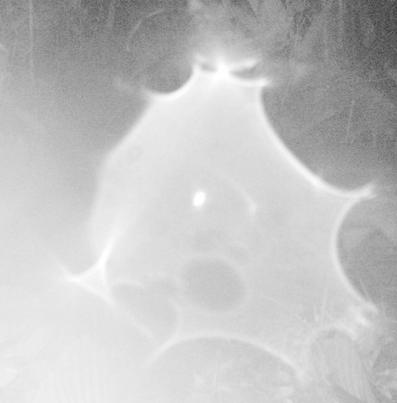}
        \caption{}
        \label{fig:unexpectedImages}
    \end{subfigure}
    \end{center}
    \caption{Different camera-trap images. (a) Ideal and scarce case in camera-trap images. (b) Over-exposed images. (c) Partial capture of the animal. (d) Different animals in the same image. (e) Blurred images. (f) Low resolution images. (g) Poor illumination. (h) Auto-occlusion. (i) Background occlusion. (j) Unexpected images. All images were taken from the Alexander von Humboldt Biological Resources Research Institute database (see section \ref{sec:DB} for more details of the database).}
    \label{fig:cameraTrapChallenges}
\end{figure}

There are many challenges in camera-trap images due to environmental conditions, animal behavior, and hardware limitation. These conditions affect the classification performance. Figure \ref{fig:cameraTrapChallenges} shows different camera-trap images after pre-processing. Image classification can be interpreted as an object recognition problem. For example Figure \ref{fig:idealImage} shows a Mazama, the problem is to recognize this image as Mazama in an automatic way. Nevertheless, images like Figure \ref{fig:idealImage} are ideal and scarce images in camera-trap framework.

There are some common problems in camera-trap images due to environmental conditions, e.g. poor illuminated images (see Figure \ref{fig:poorIllumination}), background occlusion (see Figure \ref{fig:context}), and unexpected images (see Figure \ref{fig:unexpectedImages}). Those problems can confuse algorithms that learn based on specific attributes like legs, shapes, head, coat patterns, and texture. In the same way, these images could represent a challenge even for a specialist. There are other important problems in camera-trap images due to animal behavior, e.g. different animals in the same image as Figure \ref{fig:variousAnimal} shows, and auto-occlusion as Figure \ref{fig:autoOclussion} shows. Finally, there are some problems about hardware limitation, for instance Figure \ref{fig:overExposed} shows over-exposed images, Figure \ref{fig:parts} shows partial capture of animals, Figure \ref{fig:blurred} blurred images, and Figure \ref{fig:lowResolution} shows low resolution images. The problems due to animal behavior and hardware limitations have similar consequences as environmental conditions.

\section{Identification Method}
\label{sec:identificationMethod}

\begin{figure}[H]
    \begin{center}
    \includegraphics[width=\textwidth]{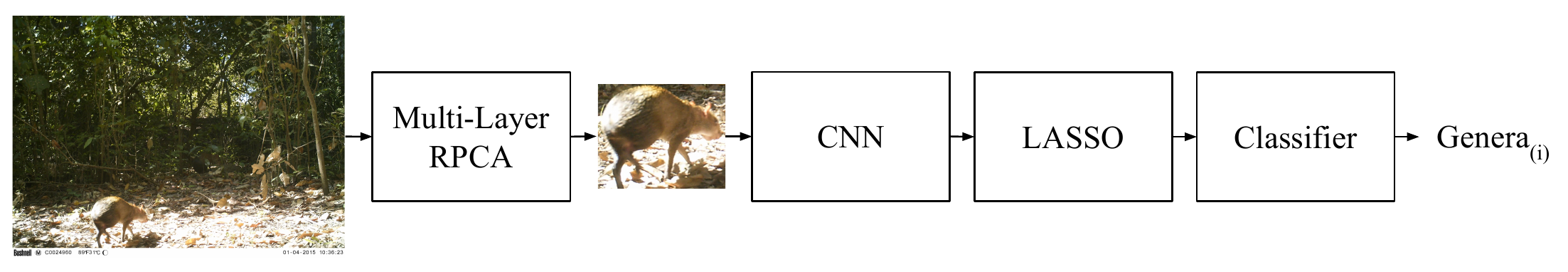}
    \end{center}
    \caption{Overall algorithm with its stages.}
    \label{fig:identificationAlgorithm}
\end{figure}

This section describes the algorithm for classifying mammal genera. Figure \ref{fig:identificationAlgorithm} shows the principal stages of our method. The segmentation stage is performed with Multi-Layer RPCA. The feature extraction and selection is achieved with CNNs and LASSO, respectively. Finally, we classified the features with ANNs or SVM.

\subsection{Multi-Layer Robust Principal Component Analysis}

Equation \ref{eqn:RPCA} shows the objective function of the RPCA, where $||L||_*$ denotes the nuclear norm of the low-rank matrix, $||S||_1$ denotes the L1-norm of the sparse matrix, and $M$ is the data matrix \cite{candes2011robust}. The Multi-Layer RPCA was proposed by Giraldo et al. \cite{giraldo2017camera}. This method combines texture and color features in the data matrix $M$, using a weight value $\beta \in [0,1]$ that indicates the contribution of the texture features to the overall data matrix. The outputs of the Multi-Layer RPCA are segmented regions, but we need the Regions Of Interest (ROI) to feed the CNNs. We merge the ROI in the Multi-Layer RPCA outputs, if exists an interception strictly greater than zero between 2 separate ROI. In this work, we fed the Multi-Layer RPCA with all the images generated by a camera (see section \ref{sec:DB} for further details), and we chose $\beta = 0.45$ according to the optimal value found in \cite{giraldo2017camera}.

\begin{align}
    \nonumber
    \text{minimize} & \quad ||L||_* + \lambda ||S||_1 \\
    \text{subject to} & \quad L+S=M
    \label{eqn:RPCA}
\end{align}



\subsection{Convolutional Neural Networks}

LeCun et al. \cite{lecun1998gradient} proposed the CNNs in 1998. These are a feed-forward artificial neural network inspired by the animal visual cortex, and consisting of multiple layer of convolutions and sub-samplings (pooling). Usually, CNNs end with an inner product and a softmax layer for assigning probabilities to each class. There are several mainstream deep CNNs architectures, in this work the following are used: GoogLeNet \cite{szegedy2015going}, ResNet50, ResNet101, and ResNet152 \cite{he2016deep}.

A CNN can be used as a black-box feature extractor, fine-tuned (transfer learning) or trained from scratch. In black-box feature extractor mode, the features (vector of features) are extracted from a pre-trained version of CNN and used to feed a classifier. Fine-tuning pretends to use a pre-trained architecture and run the back-propagation algorithm again over one or more layers in order to use previous knowledge as a seed for our data. Finally, to train from scratch implies train the CNN in the data from random weights. In this work, we fine-tune the last layer of the networks and also use them in a black-box matter. All the architectures were pre-trained in the Imagenet dataset \cite{russakovsky2015imagenet}.



\subsection{Least Absolute Shrinkage and Selection Operator}

Tibshirani \cite{tibshirani1996regression} proposed the Least Absolute Shrinkage and Selection Operator (LASSO) in 1996. LASSO is a regression analysis algorithm, that performs subset selection and regularization. Equation \ref{eqn:lasso} shows the objective function of LASSO, where $\beta_j$ are the unknown coefficients, $\textbf{x}$ is the feature matrix, $\textbf{y}$ is the output vector, and $t$ is a constraint parameter of regularization. Equation \ref{eqn:lasso} is pretty similar to \textit{least-squares} and \textit{ridge regression}. Nevertheless, LASSO has not a closed form solution due to the fact that the ridge L1 penalty makes the solution nonlinear. The L1 penalty allows some $\beta_j$ to be exactly zero, and use the LASSO as subset selection. Indeed, we used this technique as feature selector in our algorithm.

\begin{align}
\nonumber
\hat{\beta}^{lasso} = & \argmin_{\beta} \sum_{i=1}^N \left( y_i - \beta_0 - \sum_{j=1}^p x_{ij}\beta_j \right)^2 \\ 
& \text{subject to} \sum_{j=1}^p |\beta_j| \leq t
\label{eqn:lasso}
\end{align}

\subsection{Classifiers}

We performed the classification with ANNs and soft margin SVM. In this work, the soft margin SVM has a margin parameter $C$ with radial basis kernel with parameter $\gamma$. The parameters $C$ and $\gamma$ were optimized in an exhaustive search up to powers of ten, with  $10^{-2} \leq C \leq 10^{4}$ and $10^{-2} \leq \gamma \leq 10^{3}$. By comparison, the ANN has three hidden layers with sigmoid activation function, and it ends in a softmax layer. The ANN has $\tau$ number of neurons in each hidden layer, we search the best performance with $\tau=1,2,3,...,100$. The classification was evaluated with the accuracy metric in the results (see Section \ref{sec:results}).

\section{Experimental Framework}
\label{sec:expFram}

This section introduces the database used in this paper, the executed experiments, and the implementation details of our algorithm.

\subsection{Database}
\label{sec:DB}

The Alexander von Humboldt Institute performs sampling with camera traps in different regions of the Colombian forest. The database uses $176$ camera traps from $10$ different regions in Colombia. Each camera was placed in its place for $1$ to $3$ months. The images are in daytime color and nighttime infrared formats. The mammal genera of each image were labeled by biologists of the Alexander von Humboldt Institute. The database was pre-processed by experts, cutting out the animals present on the images. Likewise, we used the Multi-Layer RPCA for segmenting the animals in an automatic way. From the automatic-segmented images, we selected those with an Interception of Union (IoU) greater than 50\%, like in \cite{zhang2016animal}. Equation \ref{eqn:IoU} depicts the definition of IoU, where $A_{pred}$ is the area of the predicted region or Automatic-Segmented (AS) region, and $A_{gt}$ is the area of the Ground-Truth (GT) or expert-segmented region. Table \ref{tbl:preprocessImages} shows the number of images after the two pre-process, i.e. GT and AS region. In addition, we have 22766 False Positive (FP) regions. A FP region is a segment where $IoU \leq 0.5$, including $IoU = 0$. The manual segmented images are available upon request.

\begin{equation}
    IoU = \frac{A_{pred} \cap A_{gt}}{A_{pred} \cup	A_{gt}}
    \label{eqn:IoU}
\end{equation}

\begin{table}[h]
\begin{center}
\begin{tabular}{cccccc}
\hline
\textbf{Genera} & \textbf{\# GT} & \textbf{\# AS} & \textbf{Genera} & \textbf{\# GT} & \textbf{\# AS} \\ \hline
Mazama     & 441  & 292  & Didelphis    & 688  & 207    \\
Pecari     & 712  & 343  & Tamandua     & 204  & 125    \\ 
Cerdocyon  & 288  & 167  & Cuniculus    & 1150 & 883    \\ 
Leopardus  & 284  & 207  & Dasyprocta   & 4228 & 3396   \\ 
Dasypus    & 741  & 389  & Proechimys   & 472  & 229    \\ \hline
\end{tabular}
\end{center}
\caption{Mammals genera after the two pre-process: Number of Automatic-Segmented and Ground-Truth images.}
\label{tbl:preprocessImages}
\end{table}


\subsection{Experiments}

The first experiment deals with the problem where we have a perfect segmentation (expert segmented regions) without any FP region. The second one confronts the problem with perfect segmentation and FP regions. The third experiment faces the automatic segmented images (only the images with $IoU > 0.5$) and FP regions. The whole experiments avoid the unbalanced nature of the database. All images were pre-processed with a Contrast Limited Adaptive Histogram Equalization (CLAHE) before feeding the CNNs. Table \ref{tbl:speciesLabels} shows the genera used in Experiments 1 and 2. Besides, Experiment 3 uses all genera in Table \ref{tbl:speciesLabels} except those written in bold.

\begin{table}[h]
\begin{center}
\begin{tabular}{cccc}
\hline
\textbf{Label} & \textbf{Genera}    & \textbf{Label} & \textbf{Genera}    \\ \hline
Ma    & Mazama             & Di     & Didelphis  \\
Pe    & Pecari             & Ta     & \textbf{Tamandua}   \\
Ce    & \textbf{Cerdocyon} & Cu     & Cuniculus  \\
Le    & Leopardus          & Das    & Dasyprocta \\
Da    & Dasypus            & Pr     & Proechimys \\ \hline
\end{tabular}
\end{center}
\caption{Group of species in the experiments.}
\label{tbl:speciesLabels}
\end{table}

The first experiment only uses the GT images. We trained the CNNs GoogLeNet, ResNet50, ResNet101, and ResNet152; taking 70\% of the images for training and 30\% for testing. We fine-tuned the last layer of the networks, using the Model Zoo of Caffe \cite{jia2014caffe}. With the CNN trained, we extract the features of the last pooling layer of each CNN for all images. Afterwards, we concatenated the features extracted of our four trained CNNs. This mixture is named MixtureNet in the results. LASSO performed feature selection over the whole features: GoogLeNet, ResNet50, ResNet101, ResNet152, and MixtureNet; computing the mean squared error in LASSO with a 5-fold cross validation. Finally, we trained ANNs and SVM with the raw and the LASSO-processed features. The second experiment is pretty similar to the first one, because we use the GT images. But, we added a class of FP regions, taking 204 random images from the FP regions detected with a $IoU \leq 0.5$. In practice, we only selected regions with $IoU = 0$ as false positive for training and testing, with the aim to avoid mixing possible animal patches with FP. The third experiment use the AS images. In addition, we added 207 random images from the FP regions with $IoU = 0$. As stated before, we trained the CNNs mentioned above with finetunning. Later, we extract the features from the last pooling layers for training ANN and SVM, including the MixtureNet. The experiments were designed with the aim of comparing the results of manual and automatic segmented images.


\subsection{Implementation Details}

The CNNs were implemented in the deep learning framework Caffe \cite{jia2014caffe}. The RPCA algorithms were computed using the Sobral et al. library \cite{lrslibrary2015}. The multiclass SVM classifier was implemented with the LIBSVM library \cite{chang2011libsvm}. The rest of the source code was developed using Matlab.

\section{Results}
\label{sec:results}

This section shows the results, discussions and limitations of the experiments introduced in Section \ref{sec:expFram}.

Table \ref{tbl:resultsExperiment1} shows the results of Experiment 1. In the table, \textit{Accuracy} is the performance with the whole set of features, and \textit{Accuracy LASSO} is the performance with the features selected by LASSO. The best performance was 90.32\% of accuracy, training an ANN with the ResNet152 features. Moreover, training the classifiers (ANN and SVM) after the LASSO selection makes the results more robust against bad features, for example the accuracy of the GoogLeNet features improves after LASSO selection. Other important contribution of LASSO is the dimensional reduction, showing a sparse nature of the CNNs. Although LASSO did not improve the best result, this method accelerates the training time of the classifier (due to the dimensional reduction).

\begin{table}[h]
\begin{center}
\begin{tabular}{lcccc}
\hline
\multicolumn{1}{c}{\multirow{2}{*}{\textbf{CNN}}} & \multicolumn{2}{l}{\textbf{Accuracy [\%]}} & \multicolumn{2}{l}{\textbf{Accuracy LASSO [\%]}} \\
\multicolumn{1}{c}{}                     & \textbf{ANN}           & \textbf{SVM}          & \textbf{ANN}                & \textbf{SVM}    \\ \hline
GoogLeNet                                & 10       & 10      & 86.39       & 85.41  \\
ResNet50                                 & 88.85    & 36.88   & 85.08       & 86.23  \\
ResNet101                                & 90       & 34.75   & 87.21       & 86.39  \\
ResNet152                                & \textbf{90.32}    & 46.56   & 89.18       & 88.36  \\
MixtureNet                               & 10       & 10      & 90.16       & 88.69  \\ \hline
\end{tabular}
\end{center}
\caption{Results using expert segmentation.}
\label{tbl:resultsExperiment1}
\end{table}

Table \ref{tbl:resultsExperiment2} shows the results of Experiment 2. The best performance was 90.15\% of accuracy, training an ANN with the ResNet101 and ResNet152 features. The LASSO selection gives a robustness against bad features as in the results of Experiment 1. Finally, Table \ref{tbl:resultsExperiment3} shows the results of Experiment 3. The best performance was 92.65\% of accuracy, training an ANN with the MixtureNet features after the LASSO selection. In this experiment, LASSO selection gives robustness against bad features and the best performance. As a consequence, this method is useful for combining CNNs (trained separately) and select robust features for classification.

\begin{table}[h]
\begin{center}
\begin{tabular}{lcccc}
\hline
\multicolumn{1}{c}{\multirow{2}{*}{\textbf{CNN}}} & \multicolumn{2}{l}{\textbf{Accuracy [\%]}} & \multicolumn{2}{l}{\textbf{Accuracy LASSO [\%]}} \\
\multicolumn{1}{c}{}                     & \textbf{ANN}           & \textbf{SVM}          & \textbf{ANN}                & \textbf{SVM}    \\ \hline
GoogLeNet                                & 9.10           & 9.10         & 73.43              & 73.25  \\
ResNet50                                 & 89.25          & 42.98        & 87.46              & 87.16  \\
ResNet101                                & \textbf{90.15} & 42.39        & 88.81              & 87.16  \\
ResNet152                                & \textbf{90.15} & 34.18        & 88.36              & 86.42  \\
MixtureNet                               & 9.10          & 9.10         & 89.55               & 87.91  \\ \hline
\end{tabular}
\end{center}
\caption{Results with expert segmentation and false positive class.}
\label{tbl:resultsExperiment2}
\end{table}

\begin{table}[h]
\begin{center}
\begin{tabular}{lcccc}
\hline
\multicolumn{1}{c}{\multirow{2}{*}{\textbf{CNN}}} & \multicolumn{2}{l}{\textbf{Accuracy [\%]}} & \multicolumn{2}{l}{\textbf{Accuracy LASSO [\%]}} \\
\multicolumn{1}{c}{}                     & \textbf{ANN}           & \textbf{SVM}          & \textbf{ANN}                & \textbf{SVM}    \\ \hline
GoogLeNet                                & 11.11     & 11.11        & 90.84      & 88.87   \\
ResNet50                                 & 92.47     & 25.81        & 90.50      & 89.96   \\
ResNet101                                & 91.74     & 45.06        & 91.38      & 88.69   \\
ResNet152                                & 91.74     & 28.72        & 89.59      & 89.23   \\
MixtureNet                               & 11.11     & 11.11        & \textbf{92.65}      & 90.86   \\ \hline
\end{tabular}
\end{center}
\caption{Results with automatic segmentation and false positive class.}
\label{tbl:resultsExperiment3}
\end{table}

\begin{figure}[h]
    \begin{center}
    \begin{subfigure}[b]{0.4\textwidth}
		\centering
		\includegraphics[height=2.5cm,keepaspectratio,]{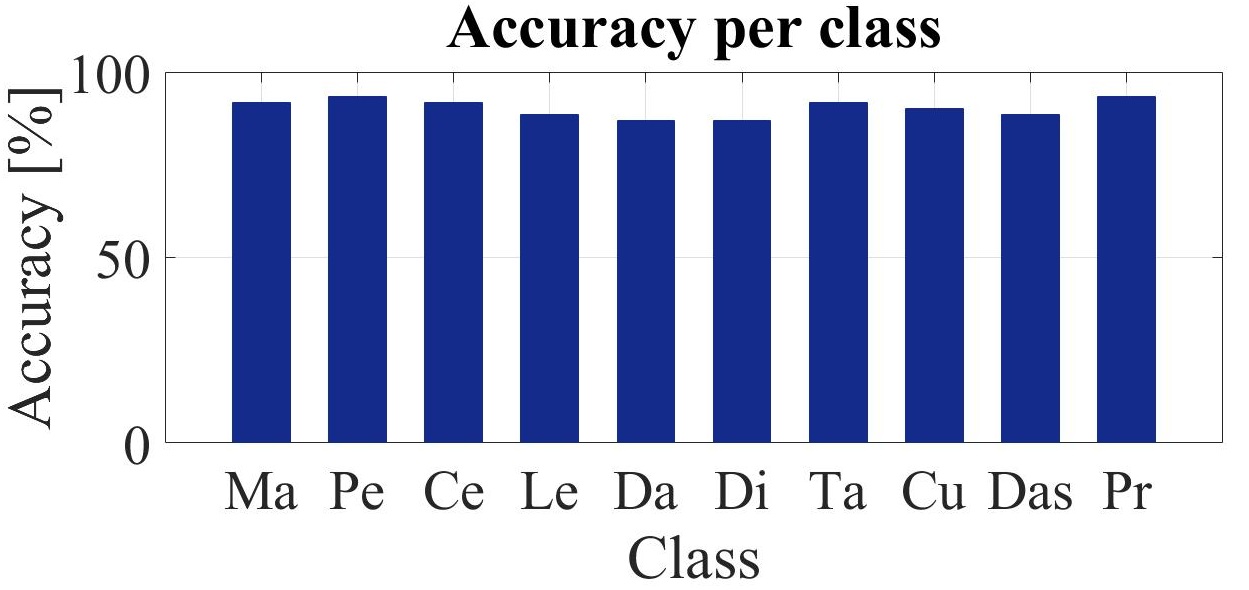}
        \caption{}
        \label{fig:accuPerClass1}
    \end{subfigure}
    \begin{subfigure}[b]{0.4\textwidth}
		\centering
		\includegraphics[height=2.5cm,keepaspectratio,]{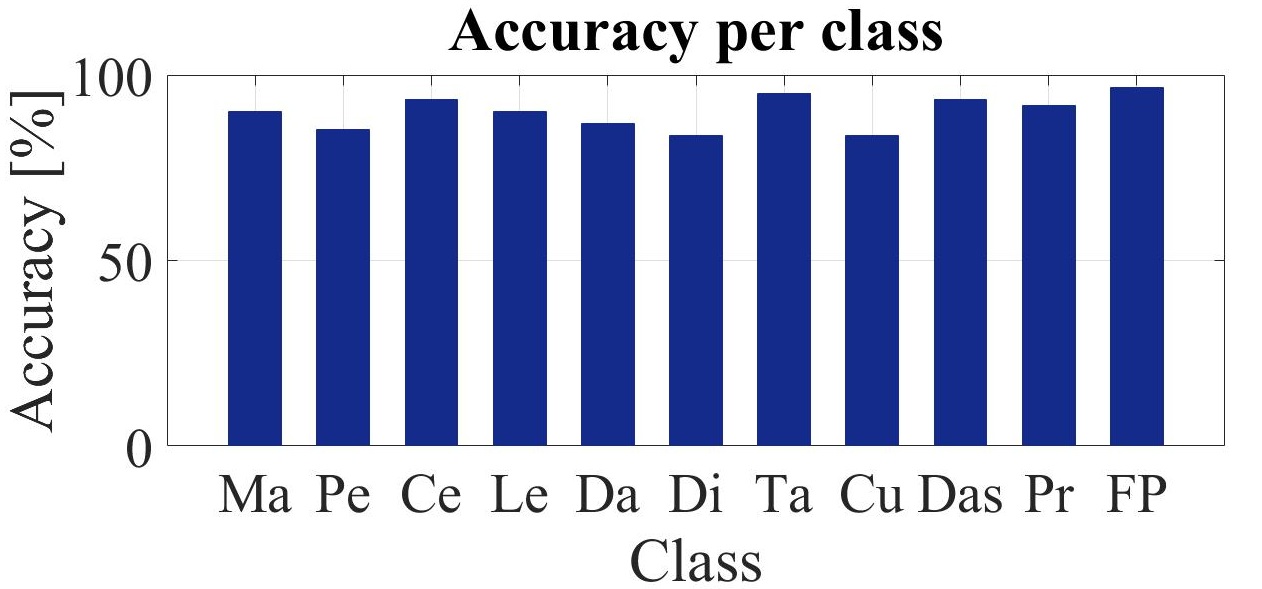}
        \caption{}
        \label{fig:accuPerClass2}
    \end{subfigure}
    \begin{subfigure}[b]{0.4\textwidth}
		\centering
		\includegraphics[height=2.5cm,keepaspectratio,]{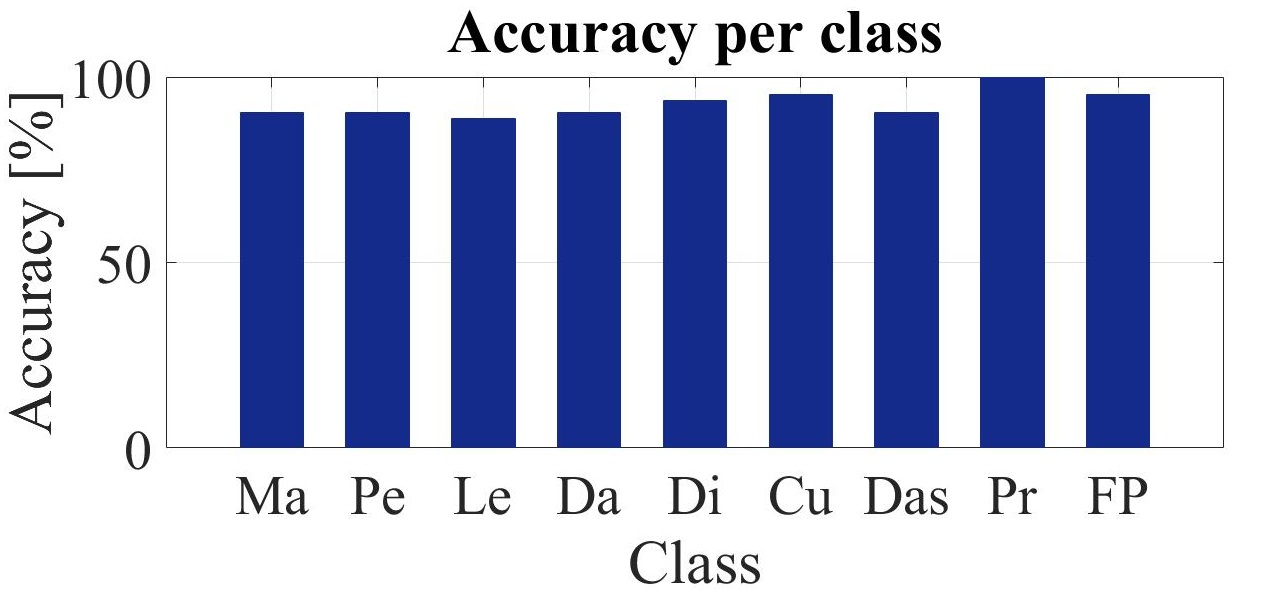}
        \caption{}
        \label{fig:accuPerClass3}
    \end{subfigure}
    \end{center}
    \caption{Accuracy per class in the experiments with the best CNN architecture in each case. (a) Accuracy per class in the experiment with expert segmentation. (b) Accuracy per class in the experiment with expert segmentation and false positive class. (c) Accuracy per class in the experiment with automatic segmentation and false positive class.}
    \label{fig:accuracyPerClass}
\end{figure}

Figure \ref{fig:accuracyPerClass} shows the accuracy per class with the architectures of the best performances in Tables \ref{tbl:resultsExperiment1}, \ref{tbl:resultsExperiment2}, and \ref{tbl:resultsExperiment3}. The standard deviation intra-class is 2.50\%, 4.61\%, and 3.62\% for Experiments 1, 2, and 3, respectively. These results offer stable accuracies in each class. Although the Multi-Layer RPCA generates thousands of false positives, the CNNs can handle this problem as shown by the results of FP in Experiments 2 and 3 (see Figure \ref{fig:accuracyPerClass}). Correspondingly, the performances suggest that our method can perform the segmentation and classification of mammal genera from camera-trap images.



We reveal the sparsity nature of the CNNs tested in our problem \cite{liu2015sparse}. Table \ref{tbl:sparsity} shows the sparsity of each pooling layer of the CNNs tested in this paper. The LASSO selection zeros out more than 96\% of features in the MixtureNet. In the same way, LASSO selection makes robust the features against noisy-features, and it accelerates the training of the ANN and SVM classifiers.


\begin{table}[h]
\begin{center}
\begin{tabular}{lccc}
\hline
\multicolumn{1}{c}{\multirow{2}{*}{\textbf{CNN}}} & \multicolumn{3}{c}{\textbf{Sparsity {[}\%{]}}}      \\
\multicolumn{1}{c}{}                     & \textbf{Experiment 1} & \textbf{Experiment 2} & \textbf{Experiment 3} \\ \hline
GoogLeNet                                & 77.15        & 79.69        & 84.28        \\
ResNet50                                 & 94.82        & 93.46        & 88.43        \\
ResNet101                                & 92.14        & 89.99        & 90.09        \\
ResNet152                                & 89.16        & 89.31        & 90.97        \\ 
MixtureNet                               & \textbf{96.71}        & 96.48        & 96.65        \\ \hline
\end{tabular}
\end{center}
\caption{Sparsity of features with LASSO selection.}
\label{tbl:sparsity}
\end{table}

\subsection{Limitations}

Our method only tested images with $IoU > 0.5$ and $IoU = 0$ for animals and FP, respectively, in Experiment 3. Certainly, regions with $0 < IoU \leq 0.5$ are limitations of our method. Nevertheless, these regions can confuse the learning algorithm, training the CNNs with unrepresentative-animal patches. Furthermore, regions with $0 < IoU \leq 0.5$ are likely false detection of the Multi-Layer RPCA.


\section{Conclusions}
\label{sec:conclusions}

We proposed an automatic genera recognition method based on Multi-Layer RPCA and deep CNNs. The proposed algorithm was composed of Multi-Layer RPCA segmentation, CNN feature extraction, LASSO features selection, and ANN or SVM classification. We tested our algorithm in camera-trap images with very challenging conditions such as environmental conditions and hardware limitation. We reached an accuracy of 92.65\% for 9 classes (including a class of False Positive regions) with automatic-segmented images, these images have an $IoU > 0.5$ with respect to the ground-truth segmentation. In addition, we reached accuracies of 90.32\% and 90.15\% for 10 and 11 classes with expert-segmented images. The LASSO selection demonstrated the sparsity of the CNNs, making zero more than 96\% of features in the MixtureNet. The results showed that CNNs can classify the automatic-segmented regions, using patches with $IoU > 0.5$.


For future work, it is important to study the performance of the method when we feed the CNNs with all outputs of the Multi-Layer RPCA (images with $IoU \geq 0$). In the same way, the unbalanced problem should be solved, this problem could be approached with one-shot learning \cite{santoro2016one}, or modifications to the stochastic gradient descent algorithm in the CNNs. Another task is the use of some Sparse Convolutional Neural Networks (SCNN) \cite{liu2015sparse} and design them based on LASSO selection.
\\
\textbf{Acknowledgment.} This work was supported by the Colombian National Fund for Science, Technology and Innovation, Francisco Jos\'e de Caldas - COLCIENCIAS (Colombia). Project No. 111571451061.

\bibliography{egbib}
\end{document}